\documentclass[runningheads]{llncs}
\usepackage{graphicx}
\usepackage{array}
\newcolumntype{P}[1]{>{\centering\arraybackslash}p{#1}}

\begin{document}
\title{Training Deep Learning Models via Synthetic Data: Application in Unmanned Aerial Vehicles}
%

\author{Andreas Kamilaris\inst{1}\inst{2}, Corjan van den Brink\inst{1} and Savvas Karatsiolis\inst{3}}


%
\authorrunning{A. Kamilaris}
\titlerunning{Training DL via Synthetic Data in UAV Applications}
%
\institute{Pervasive Systems Group, Department of Computer Science\\
University of Twente, The Netherlands \\
\email{Email: a.kamilaris@utwente.nl, g.c.vandenbrink@student.utwente.nl}\\
 \url{https://www.utwente.nl/en/eemcs/ps/}
\and
Research Centre on Interactive Media, Smart Systems and Emerging Technologies (RISE), Nicosia, Cyprus\\
\email{Email: a.kamilaris@rise.org.cy}\\ 
\url{http://www.rise.org.cy/}
\and
Department of Computer Science, University of Cyprus, Nicosia, Cyprus \\
\email{Email: skarat01@cs.ucy.ac.cy}\\
 \url{https://www.cs.ucy.ac.cy}
}

\maketitle              

\begin{abstract}
This paper describes preliminary work in the recent promising approach of generating synthetic training data for facilitating
the learning procedure of deep learning (DL) models, with a focus on aerial photos produced by unmanned aerial vehicles (UAV). The general concept and methodology are described, and
preliminary results are presented, based on a classification problem of fire identification in forests as well as a counting problem of estimating number of houses in urban areas. The proposed technique
constitutes a new possibility for the DL community, especially related to UAV-based imagery analysis, with much potential, promising results, and unexplored ground for further research.

\keywords{UAV  \and Deep Learning \and Generative Data \and Aerial Imagery}
\end{abstract}

\section{Introduction}

Deep learning (DL) constitutes a recent, modern technique for image processing and data analysis with
large potential \cite{schmidhuber2015deep}. DL belongs to the machine learning (ML) computational field and is similar to artificial neural networks (ANN). DL extends ML by adding more "depth" (complexity) into the model,
transforming the data using various functions that allow data representation in a hierarchical way, through
several abstraction levels.
DL seems to be offering better precision results in classification and/or counting computer vision-related problems, in comparison to traditional techniques such as Scalable Vector Machines and Random Forests, according to relevant surveys \cite{kamilaris2018deep}. 

An advantage of DL is the reduced need of feature engineering (FE). Previously, traditional approaches
for image classification were based on hand-engineered features, whose performance affected the
results heavily \cite{amara2017deep}. Although DL does not require FE, it still needs appropriate datasets as input in DL models during learning. These datasets need to be large, to allow DL models to learn the problem elaborately, and
expressive, to capture the variation of classes/features that need to be classified/predicted at the model
output. An existing problem is the limited availability of such appropriate datasets. This limitation makes
DL models sometimes difficult to generalize and to learn the problem well, towards high precision.

Towards addressing this limitation, a recent possibility is the generation of synthetic datasets to train DL models \cite{KamilarisSept2018simDeepL}, \cite{kar2019meta}, \cite{puig2018virtualhome}. Models are trained using synthetic images, and they are then able to classify images of the real world, or count objects encountered in the real-world images, via this transfer learning-based method.

The contribution of this paper is twofold: on one hand, to present state of art research in generating synthetic data for training DL models. On the other hand, to present preliminary work on a classification problem of fire identification in forests and a counting problem of estimating number of houses in urban areas, based on two datasets comprised of aerial photos.

\section{Related Work}
\label{RelWork}
DL is divided in discriminative and generative models \cite{goodfellow2014generative}. The
former is about predictions/classifications, and the latter about synthesis/generation of data similar to the
input datasets. The use of generative data to train DL models is promising, with
early attempts in agriculture indicating positive outcomes \cite{kamilaris2018deep}. 

Table \ref{tab1} lists related work in the field of generating training data to train DL models. The year of publication for every paper reveals how modern this technique is. Please note that we avoided adding details about performance metrics and evaluation results for each paper, because each author used different metrics and experimented on different real-world datasets for testing. However, the general conclusion in all papers was that the performance according to the metric(s) used, was better than baseline (i.e. datasets not enhanced with synthetic data) or state-of-art related work.

From Table \ref{tab1}, it is evident that related work has not entered yet the domain of UAV-based imagery analysis. The only exception is Meta-Sim \cite{kar2019meta}, which tries to learn a generative model of synthetic scenes automatically, via probabilistic scene grammars, and then it obtains images and their corresponding ground-truth via a graphics engine. Meta-Sim validates this idea addressing the problem of semantic segmentation of simulated aerial views of simple roadways. Beyond this work, to our knowledge, no other work has focused yet on generative data-based approaches for UAV-based imaging-related applications.

\begin{table} [htb]
\caption{Related work in generative data for training DL models.}
\label{tab1}
\centering
\begin{tabular}{| P{1.0cm} | p{10.0cm} | P{0.8cm} |}
\hline
\textbf{Year} &  \textbf{Purpose} & \textbf{Ref.} \\
\hline
2007 &  Simulating fluorescence microscope images of cell populations for automated image cytometry &  \cite{lehmussola2007computational} \\
\hline
2016 & Enhancing soil images coming from X-ray tomography, generating roots to help the model identify the roots from the soils & \cite{douarre2016deep} \\
\hline
2016 & Simulating top-down images of overlapping plants on soil background, to classify 23 different weed species and maize. & \cite{dyrmann2016pixel} \\
\hline
2016 & Generating fully labeled, dynamic, and photo-realistic proxy virtual world, with a focus on objects of interest, e.g. cars. & \cite{gaidon2016virtual} \\
\hline
2016 & Generating synthetic data for semantic segmentation of outdoor scenes, for recognizing aspects such as roads, buildings, cars, people, lights etc. & \cite{richter2016playing} \\
\hline
2016 & Automatically generating realistic synthetic images with pixel-level annotations for semantic segmentation & \cite{ros2016synthia} \\
\hline
2017 & Creating synthetic images to predict number of tomatoes in the images. & \cite{rahnemoonfar2017deep} \\
\hline
2018 & Generating synthetic data to identify melanoma skin cancer. & \cite{KamilarisSept2018simDeepL} \\
\hline
2018 & Synthetic data for 2D bounding box car detection. & \cite{prakash2018structured} \\
\hline
2018 & Generating 3D scenes of visually realistic houses, ranging from single-room studios to multi-storied houses, equipped with a diverse set of fully labeled 3D objects, textures and scene layout, for teaching an agent to navigate in an unseen 3D environment. & \cite{wu2018building} \\
\hline
2018 & Generating scenes for teaching an artificial agent to execute tasks in a simulated household environment. & \cite{puig2018virtualhome} \\
\hline
2019 & a) Generating data for semantic segmentation of aerial views of roadways. b) Simulating urban scenes for object detection in urban car driving. & \cite{kar2019meta} \\
\hline
\end{tabular}
\end{table}

\section{Methodology}
\begin{figure} [htb]
\centering
\includegraphics[width=1.0\textwidth]{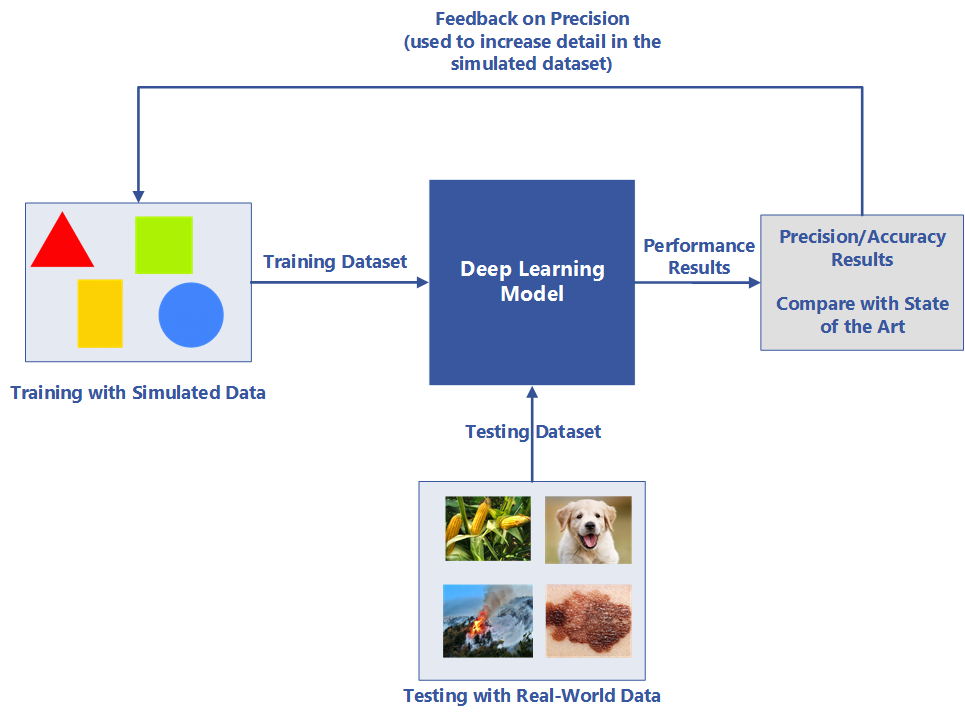}
\caption{Basic methodology in generating data for training DL models.} 
\label{fig1}
\end{figure}

The general methodology followed in this paper is illustrated in Figure \ref{fig1}. More advanced and recent proposals based on this general methodology will be discussed in Section \ref{disc}. DL models are trained with synthetic
data, and then tested with real-world data. The precision/accuracy results are analyzed and compared with the state-of-art related work (if available), and the observations made are given as feedback to the creation process of the synthetic datasets, to become more detailed and complete (e.g. to include some aspects of the real-world data not included originally, but which affect the model's prediction capabilities).

Our approach in synthetic dataset design would be to understand how DL models perform classification,
based on the existing real-world datasets (i.e. problem under study for classification or counting). 
To achieve this, we take advantage of the work in \cite{olah2018building}, which allows to visualize what happens inside DL models, i.e. which aspects/characteristics of the image are the ones that trigger the final classification. These characteristics could then be used to better design the synthetic datasets, emphasizing on these aspects when creating the simulated images.
In this paper, we focused on two different applications:
\begin{enumerate}
 \item A classification problem of identifying fires in forest areas from aerial photos.
 \item A counting problem of estimating number of houses from aerial photos.
\end{enumerate}

The former is useful for UAV which monitor forest areas for fires and smoke, while the latter would be useful for policy-makers who want to understand distribution of houses in urban areas, possibilities for photovoltaic systems, urban gardening in roofs etc.

\begin{figure*} [htb]
\begin{minipage}{.24\textwidth}
  \centering
  \includegraphics[width=1.00\linewidth]{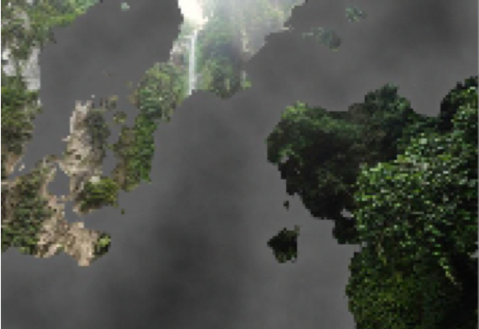}
\end{minipage}%
\begin{minipage}{.24\textwidth}
  \centering
  \includegraphics[width=0.90\linewidth]{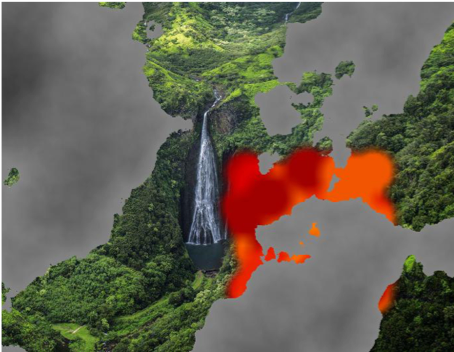}
  \end{minipage}
\begin{minipage}{.24\textwidth}
  \centering
  \includegraphics[width=0.80\linewidth]{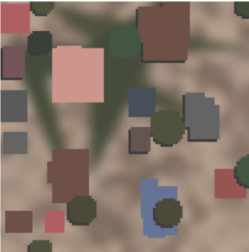}
\end{minipage}
\begin{minipage}{.24\textwidth}
  \centering
  \includegraphics[width=0.80\linewidth]{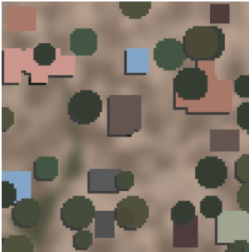}
\end{minipage}
\newline
\vspace{0.1cm}
\newline
\begin{minipage}{.24\textwidth}
  \centering
  \includegraphics[width=0.95\linewidth]{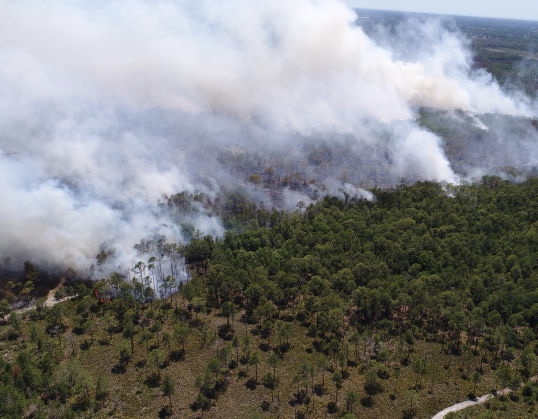}
\end{minipage}%
\begin{minipage}{.24\textwidth}
  \centering
  \includegraphics[width=0.88\linewidth]{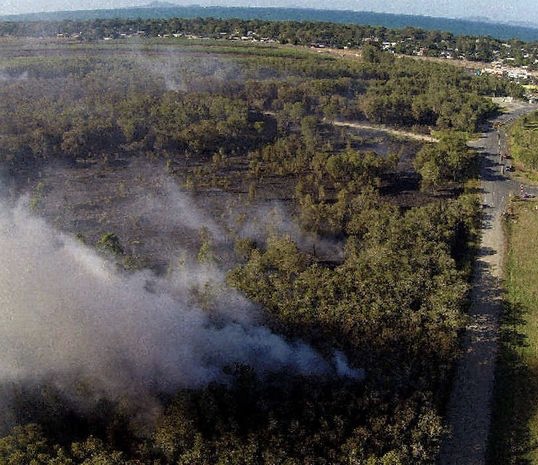}
  \end{minipage}
\begin{minipage}{.24\textwidth}
  \centering
  \includegraphics[width=0.85\linewidth]{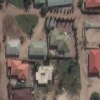}
\end{minipage}
\begin{minipage}{.24\textwidth}
  \centering
  \includegraphics[width=0.85\linewidth]{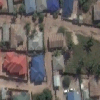}
\end{minipage}
\caption{Example images from the synthetic datasets (top). Example images from the real-world datasets (bottom). Images on the left are for the fire identification scenario, while images on the right for the case of the estimation of number of houses.}
\label{fig2}
\end{figure*}

For the problems under study, the synthetic datasets (used for training the DL model) have been created by means of Python, by using the Python Imaging Library\footnote{Python Imaging Library. https://pypi.python.org/pypi/PIL} and
OpenCV\footnote{OpenCV. https://pypi.python.org/pypi/opencv-python}. PIL libraries allow to combine graphics creation, together with programming code and computer logic, using code in order to create dots, lines, rectangles, polygons, circles, ellipses and combinations, allowing to add color, transparency, borders and outlines, but also to include filters such as "Gaussian Blur", smoothen the image, enhance the edges etc. 
By means of Python scripts, based on the PIL graphic features, we created more complex structures such as smoke, fire, houses, trees, fences, gardens etc. Samples of the synthetic data for the scenarios under study are depicted in Figure \ref{fig2} (top).

Regarding the real-world datasets (used for testing the DL model), for the fire identification case, 100 aerial photos were downloaded from Google Images, 50 of them showing forest areas and another 50 showing a forest fire. For the counting houses case, 20 aerial photos from urban areas of Tanzania have been selected, from the Open AI Tanzania Challenge\footnote{Open  AI Tanzania Challenge. https://blog.werobotics.org/2018/08/06/welcome-to-the-open-ai-tanzania-challenge/}. We cropped these photos in 100x100 pixel images, and counted the number of houses manually at each cropped photo. The result was a dataset of 60 images, each having $[0,38]$ houses from an aerial view. Samples of the real-world datasets for the two scenarios under study are depicted in Figure \ref{fig2} (bottom). Table \ref{tab2} describes the number of images used for training and testing of the two scenarios under study.

\begin{table} [htb]
\caption{Number of images used for training and testing of the DL models.}
\label{tab2}
\centering
\begin{tabular}{| P{2.7cm} | P{2.0cm} | p{7.1cm} |}
\hline
\textbf{Scenario} &  \textbf{Purpose} & \textbf{No. of images} \\
\hline
Fire identification & Training & 2,000 synthetic images \\
\hline
Fire identification & Testing & 100 real-world aerial photos (classified as 50 images of forest and 50 images of fire) \\
\hline
Counting houses & Training & 10,000 synthetic images (labelled with exact number of houses) \\
\hline
Counting houses & Testing & 60 real-world aerial photos (labelled with exact number of houses) \\
\hline
\end{tabular}
\end{table}

As a DL model, we used the Inception-v3 convolutional neural network (CNN) architecture \cite{szegedy2016rethinking} (with some adaptations, see below), as it is one of the fastest CNN architectures available, with high accuracy \cite{canziani2016analysis}. We used the default class provided by Keras/TensorFlow during our experiments. Data augmentation was used too.
For the counting houses case, our early experiments indicated we should perform adaptations to Inception-v3, to become more optimized for counting correctly. The most important design considerations were the following:

\begin{itemize}
 \item No pre-training with other datasets (e.g. ImageNet). Filters created by ImageNet are different than the filters required for counting houses.
 \item Use of dropout (i.e. 35\%).
 \item Max pooling instead of average pooling.
 \item Use larger filters at the convolutions at the beginning (i.e. 7x7) of the CNN.
 \item Use a larger value for stride (i.e. stride=5).
 \item Use a dense layer with only one output at the end of the CNN.
 \item Use ReLu for the prediction of final outcome.
\end{itemize}

\section{Results}
\begin{figure} [htb]
\centering
\includegraphics[width=1.0\textwidth]{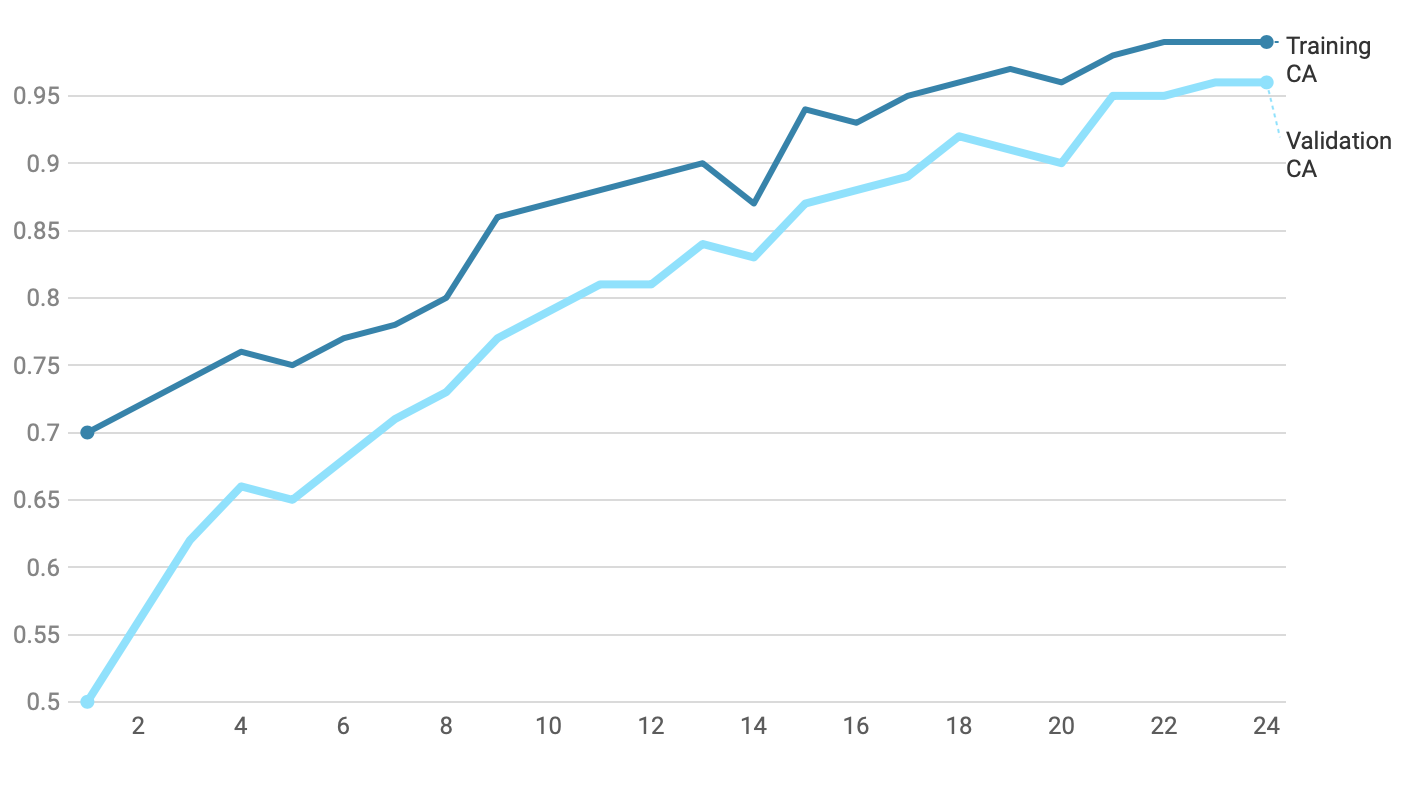}
\caption{Training and validation classification accuracy at the fire identification challenge.} 
\label{fig_fire}
\end{figure}

\begin{figure} [htb]
\centering
\includegraphics[width=1.0\textwidth]{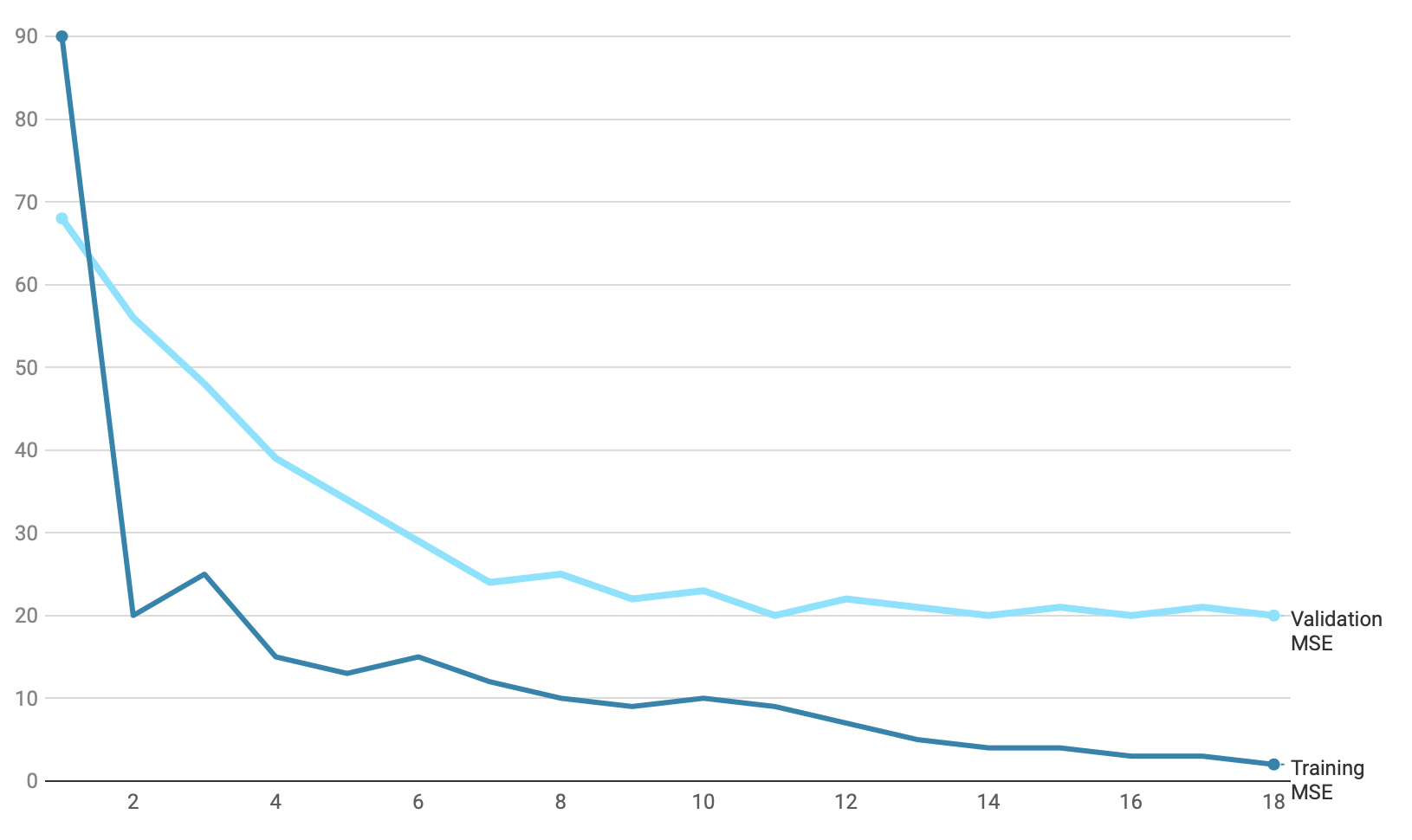}
\caption{Training and validation MSE at the counting houses challenge.} 
\label{fig_houses}
\end{figure}

Figures \ref{fig_fire} and \ref{fig_houses} show the results of the training of the DL models (i.e. synthetic data) and of the testing of the model in real-world data, for the fire identification and counting houses case respectively. Classification accuracy (CA) was used as the performance metric for the fire identification case, while Mean Square Error (MSE) for the counting houses scenario. The fire identification case required 24 epochs of training for the model to learn how to classify with $CA = 96\%$ on the validation dataset, while counting houses needed 18 epochs for the model to learn how to count with $MSE = 20$ on the validation dataset. In this case, MSE measures the average of the squares of the errors of the difference between the actual counts of houses in the images (i.e. ground truth counts of the real-world dataset) and the counts predicted by the DL model.

\section{Discussion}
\label{disc}
Results of the two scenarios under study indicate that synthetic data can prove useful for training DL models, particularly related to UAV-based aerial imagery. This evidence is backed by related work, listed in Section \ref{RelWork}. Nevertheless, we need to be cautious with these indications, because the DL models were optimized to perform well in the specific validation datasets. It is questionable (and it has not been tested) whether the DL models can produce similar results in different real-world datasets that focus on similar problems and applications \cite{kamilaris2018deep}.

The DL model for the fire identification case had very high CA. We achieved this accuracy via a hybrid approach, adding background of real forest images to the generated smoke and fire. Before this, validation CA was around 86\%.
On the other hand, the DL model for the counting problem still needs improvement. A $MSE = 20$ means that the model can predict the number of houses with an error of $\pm 4.47$ houses. For example, for a photo with 20 houses, the model would predict in the range of $[16,24]$. There is definitely space for further work on this. We note that we reached this MSE after many weeks of observations and the iterative process of adding more details to the generated synthetic dataset (see Figure \ref{fig1}). These details included trees, grass, swimming pools, fences etc.. Each of them helped to reduce the overall error.

Applications of the proposed approach can be found in various research domains and scientific disciplines, such as agriculture, life sciences, microbiology, earth sciences etc. The approach of generative data for training DL models would be extremely useful for UAV and robotics \cite{puig2018virtualhome}, \cite{wu2018building}, where computer vision is involved. It could improve operation and accuracy of automatic robots collecting crops, removing weeds or estimating yields of crops \cite{rahnemoonfar2017deep}, \cite{kamilaris2018deep}. It could also be used in disaster monitoring and surveillance \cite{KamilarisSept2017aDisMan}, where remote sensing (i.e. satellites or UAV) is used to identify events of interest (e.g. disasters, violence incidents, land cover mapping, effects on climate change etc.). Finallly, it could be used in environmental studies, e.g. to understand the environmental impact of livestock agriculture \cite{KamilarisSept2017AgriBigCat}.

Moreover, we highlight some recent state-of-art work in this domain, which relates to our proposed methodology, showing promising results in application areas other than UAV aerial imagery. First, the work in  \cite{walach2016learning} incorporates two significant improvements: layered boosting (i.e. a layered approach, where training is done in stages) and selective sampling (i.e. streamline the training process by reducing the impact of the low quality samples, such as trivial cases or outliers). Second, the concept of Structured Domain Randomization (SDR) places objects and distractors randomly according to probability distributions \cite{prakash2018structured} and from probabilistic scene grammars  \cite{kar2019meta}, which arise from the specific problem at hand. In this manner, SDR-generated imagery enables the neural network to take the context around an object into consideration during detection. Third, related specifically to counting, the work in \cite{lempitsky2010learning} evades the hard task of learning to detect and localize individual object instances. Instead, it casts the problem as that of estimating an image density whose integral over any image region gives the count of objects within that region. 
Furthermore, our work, as well as the aforementioned promising approaches \cite{walach2016learning}, \cite{prakash2018structured}, \cite{lempitsky2010learning} can be combined with Generative Adversarial Networks (GANs), to stylize synthetic images to look more like those captured in the real world \cite{li2018semantic}, \cite{zhu2017unpaired}.

Finally, we note another recent possibility, that of utilizing the \textit{Aerial Informatics and Robotics} platform \cite{shah2017aerial} for generating seamlessly training data related to UAV-based aerial imagery. This platform acts as an easy-to-use simulator that aims to enable designers and developers of robotic systems to generate graphical data. Its biggest advantage is that it uses recent advances in computation and graphics to simulate the physics and perception such that the environment realistically reflects the actual world.

\section{Conclusion}
This paper has described preliminary work in the approach of generating synthetic training data for facilitating
the learning procedure of DL models, with a focus on UAV-based aerial imagery. The general methodology of this approach was described, and preliminary results were presented, focused on two different challenges: a classification problem of fire identification in forests as well as a counting problem of estimating number of houses in urban areas. Results were promising, but there is still space for improvement, especially in the counting houses case.
Use of synthetic data for training DL models in aerial imagery is a new exciting possibility for the research community working in this area, especially in cases where ground-truth data is scarce or expensive to produce.

For future work, we aim to experiment with more realistic generation of synthetic data, by using game engines such as the Unity development platform\footnote{Unity. https://unity.com}. We also aim to apply our methodology in new UAV-related applications such as human crowd counting, identification and counting of endangered species and wild animals etc., enhancing our methodology with the new proposals described in Section \ref{disc}, in order to minimize the distribution
gap between the rendered outputs of the synthetic data and the target real-world data.

%

\bibliographystyle{splncs04}
\bibliography{uabib}

\end{document}